\documentclass{article} 
\usepackage{nips14submit_e}
\usepackage{times,graphicx}
\usepackage{url}
\usepackage{amsmath}
\usepackage{amssymb}
\usepackage[small,tight]{bibhacks}
\nipsfinalcopy

\title{Sequence to Sequence Learning\\with Neural Networks}

\author{
Ilya Sutskever \\
Google\\
\texttt{ilyasu@google.com} \\
\And
Oriol Vinyals \\
Google\\
\texttt{vinyals@google.com} \\
\And
Quoc V. Le \\
Google\\
\texttt{qvl@google.com} \\
}

\begin{document}

\maketitle

\begin{abstract}

Deep Neural Networks (DNNs) are powerful models that have achieved
excellent performance on difficult learning tasks. Although DNNs work
well whenever large labeled training sets are available, 
they cannot be used to map sequences to sequences.  In this paper, we
present a general end-to-end approach to sequence learning that makes
minimal assumptions on the sequence structure. Our method uses a
multilayered Long Short-Term Memory (LSTM) to map the input sequence
to a vector of a fixed dimensionality, and then another deep LSTM to
decode the target sequence from the vector.  Our main result is that
on an English to French translation task from the WMT'14 dataset, the
translations produced by the LSTM achieve a BLEU score of 34.8 on the
entire test set, where the LSTM's BLEU score was penalized on
out-of-vocabulary words. Additionally, the LSTM did not have
difficulty on long sentences. For comparison, a phrase-based
SMT system achieves a BLEU score of 33.3 on the same dataset.  When we
used the LSTM to rerank the 1000 hypotheses produced by the
aforementioned SMT system, its BLEU score increases to 36.5, which
is close to the previous best result on this task.  The LSTM also learned sensible
phrase and sentence representations that are sensitive to word order
and are relatively invariant to the active and the passive voice.
Finally, we found that reversing the order of the words in all source
sentences (but not target sentences) improved the LSTM's performance
markedly, because doing so introduced many short term dependencies
between the source and the target sentence which made the optimization
problem easier.

\end{abstract}

\section{Introduction}

Deep Neural Networks (DNNs) are extremely powerful machine learning
models that achieve excellent performance on difficult problems such
as speech recognition \cite{hinton12,dahl12b} and visual object
recognition \cite{kriz12,ciresan12,lecun98,le12}.  DNNs are
powerful because they can perform arbitrary parallel computation for a
modest number of steps.  A surprising example of the power of DNNs is
their ability to sort $N$ \!\!\!\!\quad $N$-bit numbers using only 2
hidden layers of quadratic size \cite{razborov}. So, while neural
networks are related to conventional statistical models, they learn an
intricate computation.  Furthermore, large DNNs can be trained with
supervised backpropagation whenever the labeled training set has
enough information to specify the network's parameters.  Thus, if
there exists a parameter setting of a large DNN that achieves good
results (for example, because humans can solve the task very rapidly),
supervised backpropagation will find these parameters and solve the
problem.

Despite their flexibility and power, DNNs can only be applied to
problems whose inputs and targets can be sensibly encoded with vectors
of fixed dimensionality.  It is a significant limitation, since many
important problems are best expressed with sequences whose lengths are
not known a-priori.  For example, speech recognition and machine
translation are sequential problems.  Likewise, question answering can
also be seen as mapping a sequence of words representing the question
to a sequence of words representing the answer.  It is therefore clear
that a domain-independent method that learns to map sequences to
sequences would be useful.

\begin{figure}[h]
\centering \includegraphics[width=0.9\textwidth]{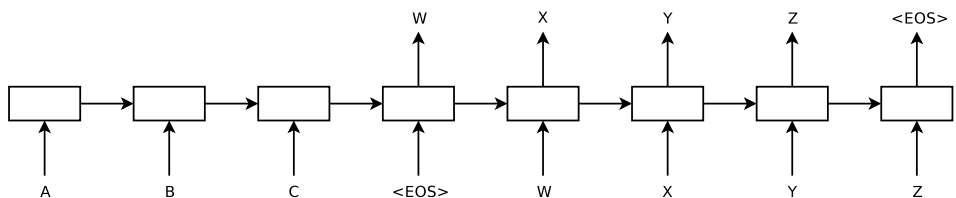}
\caption{\small Our model reads an input sentence ``ABC'' and produces
  ``WXYZ'' as the output sentence.  The model stops making predictions
  after outputting the end-of-sentence token.  Note that the LSTM
  reads the input sentence in reverse, because doing so introduces
  many short term dependencies in the data that make the optimization
  problem much easier. }
\label{fig:translation-model2}
\end{figure}

Sequences pose a challenge for DNNs because they require that the
dimensionality of the inputs and outputs is known and fixed.  
In this paper, we show that a straightforward application of the Long
Short-Term Memory (LSTM) architecture \cite{hochreiter97} can solve
general sequence to sequence problems.  The idea is to use one LSTM to
read the input sequence, one timestep at a time, to obtain large
fixed-dimensional vector representation, and then to use another LSTM
to extract the output sequence from that vector
(fig.~\ref{fig:translation-model2}).  The second LSTM is essentially a
recurrent neural network language model
\cite{rumelhart1986learning,mikolov2010recurrent,sundermeyer12} except
that it is conditioned on the input sequence.  The LSTM's ability to
successfully learn on data with long range temporal dependencies makes
it a natural choice for this application due to the considerable time
lag between the inputs and their corresponding outputs
(fig.~\ref{fig:translation-model2}).

There have been a number of related attempts to address the general
sequence to sequence learning problem with neural networks.   Our
approach is closely related to Kalchbrenner and Blunsom \cite{kal13} who were
the first to map the entire input sentence to vector, and is related 
to Cho et al.~\cite{cho14} although the latter was used only for rescoring hypotheses
produced by a phrase-based system.  Graves
\cite{graves13c} introduced a novel differentiable attention mechanism
that allows neural networks to focus on different parts
of their input, and an elegant variant of this idea was successfully
applied to machine translation by Bahdanau et al.~\cite{bog14}.  The Connectionist
Sequence Classification is another popular technique for mapping sequences
to sequences with neural networks, but it assumes a monotonic alignment
between the inputs and the outputs \cite{graves1}.


The main result of this work is the following.  On the WMT'14 English
to French translation task, we obtained a BLEU score of {\bf 34.81} by
directly extracting translations from an ensemble of 5 deep LSTMs
(with 384M parameters and 8,000 dimensional state each) using a simple left-to-right beam-search
decoder.  This is by far the best result achieved by direct
translation with large neural networks.  For comparison, the BLEU
score of an SMT baseline on
this dataset is 33.30 \cite{wmt14_en_fr}.  The
34.81 BLEU score was achieved by an LSTM with a vocabulary of 80k
words, so the score was penalized whenever the reference translation
contained a word not covered by these 80k.  This result shows that a
relatively unoptimized small-vocabulary neural network architecture which has much room
for improvement outperforms a phrase-based SMT system.

Finally, we used the LSTM to rescore the publicly available 1000-best
lists of the SMT baseline on the same task \cite{wmt14_en_fr}.  By
doing so, we obtained a BLEU score of 36.5, which improves the
baseline by 3.2 BLEU points and is close to the previous best published
result on this task (which is 37.0 \cite{durrani-EtAl:2014:W14-33}).

Surprisingly, the LSTM did not suffer on very long sentences, despite
the recent experience of other researchers with related architectures
\cite{curse}.  We were able to do well on long sentences because we
reversed the order of words in the source sentence but not the target
sentences in the training and test set. By doing so, we introduced
many short term dependencies that made the optimization problem much
simpler (see sec.~\ref{sec:model} and \ref{sec:rev_rev}).  As a result, SGD could learn
LSTMs that had no trouble with long sentences.  The simple trick of
reversing the words in the source sentence is one of the key technical
contributions of this work.
 
A useful property of the LSTM is that it learns to map an input
sentence of variable length into a fixed-dimensional vector
representation.  Given that translations tend to be paraphrases of the
source sentences, the translation objective encourages the LSTM to
find sentence representations that capture their meaning, as sentences
with similar meanings are close to each other while different
sentences meanings will be far. A qualitative evaluation supports
this claim, showing that our model is aware of word order and is
fairly invariant to the active and passive voice.

\section{The model}
\label{sec:model}

The Recurrent Neural Network (RNN) \cite{werbos,rumelhart1986learning}
is a natural generalization of feedforward neural networks to
sequences.  Given a sequence of inputs $(x_1,\ldots,x_T)$, a
standard RNN computes a sequence of outputs $(y_1,\ldots,y_T)$ by
iterating the following equation: 
\begin{eqnarray*}
h_t &=& \mathrm{sigm}\left(W^{\mathrm{hx}} x_t + W^{\mathrm{hh}} h_{t-1}\right) \\
y_t &=& W^{\mathrm{yh}}h_t
\end{eqnarray*}
The RNN can easily map sequences to sequences whenever the alignment
between the inputs the outputs is known ahead of time. However, it is
not clear how to apply an RNN to problems whose input and the output
sequences have different lengths with complicated and non-monotonic
relationships.

The simplest strategy for general sequence learning is to map the input
sequence to a fixed-sized vector using one RNN, and then to map the
vector to the target sequence with another RNN (this approach has also been
taken by Cho et al.~\cite{cho14}).  While it could work
in principle since the RNN is provided with all the relevant
information, it would be difficult to train the RNNs due to the
resulting long term dependencies
(figure \ref{fig:translation-model2})
\cite{hochreiter_long_term,bengio_long_term,hochreiter97,Hochreiter01gradientflow}. However, the Long
Short-Term Memory (LSTM) \cite{hochreiter97} is known to learn
problems with long range temporal dependencies, so an LSTM may succeed
in this setting.

The goal of the LSTM is to estimate the conditional probability
$p(y_1,\ldots,y_{T'} | x_1,\ldots,x_T)$ where $(x_1,\ldots,x_T)$ is an
input sequence and $y_1,\ldots,y_{T'}$ is its corresponding output
sequence whose length $T'$ may differ from $T$. The LSTM computes this
conditional probability by first obtaining the fixed-dimensional
representation $v$ of the input sequence $(x_1,\ldots,x_T)$ given by
the last hidden state of the LSTM, and then computing the probability
of $y_1,\ldots,y_{T'}$ with a standard LSTM-LM formulation whose
initial hidden state is set to the representation $v$ of
$x_1,\ldots,x_T$:
\begin{equation}
p(y_1,\ldots,y_{T'} | x_1,\ldots,x_T) = \prod_{t=1}^{T'} p(y_t | v, y_1, \ldots, y_{t-1})
\label{eqn:keyequation}
\end{equation}
In this equation, each $p(y_t | v, y_1, \ldots, y_{t-1})$ distribution
is represented with a softmax over all the words in the
vocabulary.  We use the LSTM formulation from Graves \cite{graves13c}.
Note that we require that each sentence ends with a special
end-of-sentence symbol ``$<$EOS$>$'', which enables the model to define a
distribution over sequences of all possible lengths. The overall
scheme is outlined in figure \ref{fig:translation-model2}, where the
shown LSTM computes the representation of ``A'', ``B'', ``C'', ``$<$EOS$>$''
and then uses this representation to compute the probability of ``W'',
``X'', ``Y'', ``Z'', ``$<$EOS$>$''.

Our actual models differ from the above description in three important
ways.  First, we used two different LSTMs: one for the input sequence
and another for the output sequence, because doing so increases the
number model parameters at negligible computational cost and makes it
natural to train the LSTM on multiple language pairs simultaneously
\cite{kal13}.  Second, we found that deep LSTMs significantly
outperformed shallow LSTMs, so we chose an LSTM with four layers.
Third, we found it extremely valuable to reverse the order of the
words of the input sentence. So for example, instead of mapping the
sentence $a,b,c$ to the sentence $\alpha, \beta, \gamma$, the LSTM is
asked to map $c,b,a$ to $\alpha, \beta, \gamma$, where $\alpha, \beta,
\gamma$ is the translation of $a,b,c$.  This way, $a$ is in close
proximity to $\alpha$, $b$ is fairly close to $\beta$, and so on, a
fact that makes it easy for SGD to ``establish communication'' between
the input and the output.  We found this simple data transformation to
greatly improve the performance of the LSTM.


\section{Experiments}
\label{sec:experiments}
 
We applied our method to the WMT'14 English to French MT task in two
ways.  We used it to directly translate the input sentence without
using a reference SMT system and we it to rescore the n-best lists of
an SMT baseline.  We report the accuracy of these translation methods,
present sample translations, and visualize the resulting sentence
representation.

\subsection{Dataset details}

We used the WMT'14 English to French dataset.  We trained our models
on a subset of 12M sentences consisting of 348M French words and 304M
English words, which is a clean ``selected'' subset from
\cite{wmt14_en_fr}. We chose this translation task and this specific
training set subset because of the public availability of a tokenized training and
test set together with 1000-best lists from the baseline SMT 
\cite{wmt14_en_fr}.

As typical neural language models rely on a vector representation for
each word, we used a fixed vocabulary for both languages.  We used
160,000 of the most frequent words for the source language and 80,000
of the most frequent words for the target language.  Every
out-of-vocabulary word was replaced with a special ``UNK'' token.  
 



\subsection{Decoding and Rescoring}

The core of our experiments involved training a large deep LSTM on
many sentence pairs. We trained it by maximizing the log
probability of a correct translation $T$ given the source sentence
$S$, so the training objective is
$$1/|\mathcal S|\sum_{(T,S)\in\mathcal S}\log p(T|S)$$ where $\mathcal
S$ is the training set.  Once training is complete, we produce
translations by finding the most likely translation according to
the LSTM:
\begin{equation}
\label{eqn:decode}
\hat{T} = \arg\max_T p(T|S)
\end{equation}
We search for the most likely translation using a simple left-to-right
beam search decoder which maintains a small number $B$ of partial
hypotheses, where a partial hypothesis is a prefix of some
translation.  At each timestep we extend each partial hypothesis in
the beam with every possible word in the vocabulary. This greatly
increases the number of the hypotheses so we discard all but the $B$
most likely hypotheses according to the model's log probability.  As soon
as the ``$<$EOS$>$'' symbol is appended to a hypothesis, it is removed from
the beam and is added to the set of complete hypotheses.  While this
decoder is approximate, it is simple to implement.  Interestingly, our
system performs well even with a beam size of 1, and a beam of
size 2 provides most of the benefits of beam search (Table
\ref{tab:blue_fr}).   


We also used the LSTM to rescore the 1000-best lists produced by the
baseline system \cite{wmt14_en_fr}.  To rescore an n-best list, we
computed the log probability of every hypothesis with our LSTM and
took an even average with their score and the LSTM's score.

\subsection{Reversing the Source Sentences}
\label{sec:rev_rev}

While the LSTM is capable of solving problems with long term
dependencies, we discovered that the LSTM learns much better when the
source sentences are reversed (the target sentences are not reversed).  By
doing so, the LSTM's test perplexity dropped from 5.8 to 4.7, and the 
test BLEU scores of its decoded translations increased from 25.9 to 30.6.

While we do not have a complete explanation to this phenomenon, we
believe that it is caused by the introduction of many short term
dependencies to the dataset.  Normally, when we concatenate a source
sentence with a target sentence, each word in the source sentence is
far from its corresponding word in the target sentence. As a result,
the problem has a large ``minimal time lag'' \cite{minimal_time_lag}.  By reversing the
words in the source sentence, the average distance between
corresponding words in the source and target language is unchanged.
However, the first few words in the source language are now very close to
the first few words in the target language, so the problem's minimal
time lag is greatly reduced. Thus, backpropagation has an easier time
``establishing communication'' between the source sentence and the
target sentence, which in turn results in substantially improved overall
 performance.

Initially, we believed that reversing the input sentences would 
only lead to more confident predictions in the early parts of the target
sentence and to less confident predictions in the later parts.
However, LSTMs trained on reversed source sentences did much better on
long sentences than LSTMs trained on the raw source sentences (see
sec.~\ref{sec:long_sentences}), which suggests that reversing the
input sentences results in LSTMs with better memory utilization.




\subsection{Training details}

We found that the LSTM models are fairly easy to train.  We used deep
LSTMs with 4 layers, with 1000 cells at each layer and 1000
dimensional word embeddings, with an input vocabulary of 160,000
and an output vocabulary of 80,000.  Thus the deep LSTM uses 8000 real 
numbers to represent a sentence. We found deep LSTMs to
significantly outperform shallow LSTMs, where each additional layer
reduced perplexity by nearly 10\%, possibly due to their much larger
hidden state.  We used a naive softmax over 80,000 words at each
output.  The resulting LSTM has 384M parameters of which 64M are pure
recurrent connections (32M for the ``encoder'' LSTM and 32M for the
``decoder'' LSTM). The complete training details are given below:
\begin{itemize}
\item We initialized all of the LSTM's parameters with the uniform distribution between
  -0.08 and 0.08
\item We used stochastic gradient descent without momentum,
  with a fixed learning rate of 0.7.  After 5 epochs, we begun
  halving the learning rate every half epoch.  We trained our models for a
  total of 7.5 epochs.
\item We used batches of 128 sequences for the gradient and divided it
  the size of the batch (namely, 128).
\item Although LSTMs tend to not suffer from the vanishing gradient
  problem, they can have exploding gradients.  Thus we enforced a hard
  constraint on the norm of the gradient
  \cite{graves13c,razvan} by scaling it when its norm exceeded
  a threshold. For each training batch, we compute $s =
  \left\|g\right\|_2$, where $g$ is the gradient divided by 128. If $s > 5$, we set
  $g = \frac{5g}{s}$.
\item Different sentences have different lengths.  Most sentences are
  short (e.g., length 20-30) but some sentences are long (e.g., length
  $>$ 100), so a minibatch of 128 randomly chosen training sentences
  will have many short sentences and few long sentences, and as a
  result, much of the computation in the minibatch is wasted.  To
  address this problem, we made sure that all sentences in a
  minibatch are roughly of the same length, yielding a 2x speedup.
\end{itemize}

\subsection{Parallelization}

A C++ implementation of deep LSTM with the configuration from the
previous section on a single GPU processes a speed of approximately
1,700 words per second.  This was too slow for our purposes, so we 
parallelized our model using an 8-GPU machine.  Each layer of the LSTM
was executed on a different GPU and communicated its activations
to the next GPU / layer as soon as they were computed.  Our models
have 4 layers of LSTMs, each of which resides on a separate GPU.  The remaining
4 GPUs were used to parallelize the softmax, so each GPU was
responsible for multiplying by a $1000\times 20000$ matrix.  The
resulting implementation achieved a speed of 6,300 (both English and
French) words per second with a minibatch size of 128. 
Training took about a ten days with this implementation.

\subsection{Experimental Results}

We used the cased BLEU score \cite{bleu} to evaluate the quality of our
translations. We computed our BLEU scores using \texttt{multi-bleu.pl}\footnote{
There several variants of the BLEU score, and each variant is defined  with a perl script. } 
on the \emph{tokenized} predictions and ground truth.
This way of evaluating the BELU score is consistent with \cite{cho14} and \cite{bog14}, and reproduces
the 33.3 score of \cite{wmt14_en_fr}.
However, if we evaluate the best WMT'14 system \cite{durrani-EtAl:2014:W14-33}
(whose predictions can be downloaded from \url{statmt.org\matrix}) in this manner, we get   
37.0, which is greater than the 35.8 reported by \url{statmt.org\matrix}.  



The results are presented in tables \ref{tab:blue_fr} and
\ref{tab:blue_fr_rescore}.  Our best results are obtained with an
ensemble of LSTMs that differ in their random initializations and
in the random order of minibatches.  While the decoded translations of the
LSTM ensemble do not outperform the best WMT'14 system, it is the first time
that a pure neural translation system outperforms a 
phrase-based SMT baseline on a large scale MT task by a sizeable margin,
despite its inability to handle out-of-vocabulary words.  The LSTM
is within 0.5 BLEU points of the best WMT'14 result if it is used to rescore the 1000-best
list of the baseline system.

\begin{table}[t]
\centering
\begin{small}
\begin{tabular}{|c|c|}
\hline
{\bf Method}  & {\bf test BLEU score (ntst14) } \\ \hline
Bahdanau et al. \cite{bog14}  &  28.45 \\ \hline
Baseline System  \cite{wmt14_en_fr} & 33.30 \\ \hline
\hline
Single forward LSTM, beam size 12 & 26.17 \\ \hline                 
Single reversed LSTM, beam size 12 & 30.59 \\ \hline
Ensemble of 5 reversed LSTMs, beam size 1  &  33.00 \\ \hline
Ensemble of 2 reversed LSTMs, beam size 12  &  33.27 \\ \hline
Ensemble of 5 reversed LSTMs, beam size 2  &  34.50 \\ \hline
Ensemble of 5 reversed LSTMs, beam size 12  &  {\bf 34.81} \\ \hline
\end{tabular}
\end{small}
\caption{The performance of the LSTM on WMT'14 English to French test
  set (ntst14).  Note that an ensemble of 5 LSTMs with a beam of size
  2 is cheaper than of a single LSTM with a beam of size 12.  }
\label{tab:blue_fr}
\end{table}

\begin{table}[]
\centering
\begin{small}
\begin{tabular}{|c|c|}
\hline
{\bf Method}  & {\bf test BLEU score (ntst14) } \\ \hline
Baseline System  \cite{wmt14_en_fr} & 33.30 \\ \hline
Cho et al. \cite{cho14}  & 34.54 \\ \hline 
Best WMT'14 result \cite{durrani-EtAl:2014:W14-33} &  {\bf 37.0} \\ \hline
\hline
Rescoring the baseline 1000-best with a single forward LSTM & 35.61 \\ \hline 
Rescoring the baseline 1000-best with a single reversed  LSTM & 35.85 \\ \hline  
Rescoring the baseline 1000-best with an ensemble of 5 reversed LSTMs  &  {\bf 36.5} \\ \hline    
\hline
Oracle Rescoring of the Baseline 1000-best lists    & $\sim$45 \\ \hline 
\end{tabular}
\end{small}
\caption{Methods that use neural networks together with an SMT system
  on the WMT'14 English to French test set (ntst14).}
\label{tab:blue_fr_rescore}
\end{table}

\subsection{Performance on long sentences}
\label{sec:long_sentences}

\begin{figure}[h!]
\centerline{
\includegraphics[width=1.\textwidth]{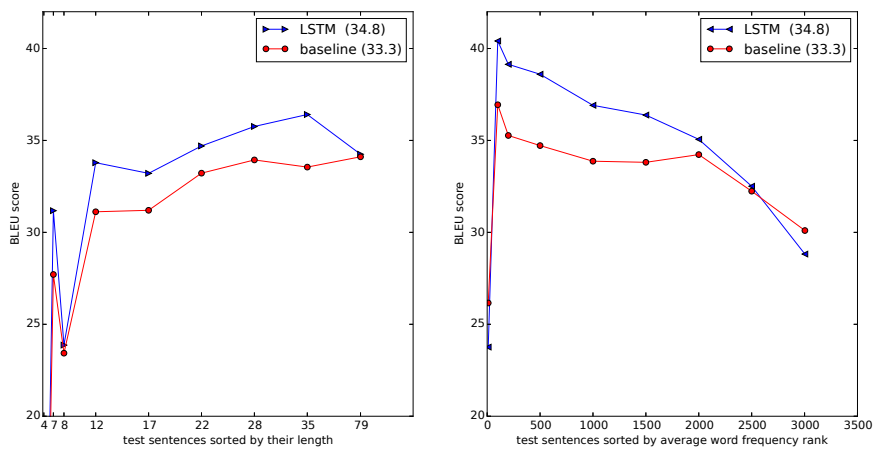} }  
\caption{\small The left plot shows the performance of our system as a
  function of sentence length, where the x-axis corresponds to the
  test sentences sorted by their length and is marked by the actual
  sequence lengths.  There is no degradation on sentences with less
  than 35 words, there is only a minor degradation on the longest
  sentences.  The right plot shows the LSTM's performance on sentences
  with progressively more rare words, where the x-axis corresponds to
  the test sentences sorted by their ``average word frequency rank''.
}
\label{fig:oriol}
\end{figure}

We were surprised to discover that the LSTM did well on long
sentences, which is shown quantitatively in figure \ref{fig:oriol}.
Table \ref{tab:examples} presents several examples of long sentences and
their translations. 

\begin{table}[ht!]
\centering
\begin{footnotesize}
\begin{tabular}{|l|l|}
\hline
{\bf  Type} & {\bf Sentence} \\ 
\hline
\hline
{\bf Our model} & Ulrich UNK , membre du conseil d' administration du constructeur automobile Audi , \\
& affirme qu' il s' agit d' une pratique courante depuis des ann\'{e}es pour que les t\'{e}l\'{e}phones \\
& portables  puissent  \^{e}tre collect\'{e}s avant les r\'{e}unions du conseil d' administration afin qu' ils\\
&  ne soient pas  utilis\'{e}s comme appareils d' \'{e}coute \`{a} distance .\\
\hline
{\bf  Truth} &  Ulrich Hackenberg , membre du conseil d' administration du constructeur automobile Audi ,   \\
& d\'{e}clare que la collecte des t\'{e}l\'{e}phones portables avant les r\'{e}unions du conseil , afin qu' ils  \\
& ne puissent pas \^{e}tre utilis\'{e}s comme appareils d' \'{e}coute \`{a} distance , est une pratique courante \\ 
& depuis des ann\'{e}es .\\
\hline\hline
{\bf Our model} & 
`` Les t\'{e}l\'{e}phones cellulaires , qui sont vraiment une question , non seulement parce qu' ils \\
& pourraient potentiellement causer des interf\'{e}rences avec les appareils de navigation , mais \\
& nous savons , selon la FCC , qu' ils pourraient interf\'{e}rer avec les tours de t\'{e}l\'{e}phone cellulaire \\
& lorsqu' ils sont dans l' air '' , dit UNK .\\
\hline
{\bf Truth} & 
`` Les t\'{e}l\'{e}phones portables sont v\'{e}ritablement un probl\`{e}me , non seulement parce qu' ils \\
& pourraient \'{e}ventuellement cr\'{e}er des interf\'{e}rences avec les instruments de navigation , mais \\
& parce que nous savons , d' apr\`{e}s la FCC , qu' ils pourraient perturber les antennes-relais de \\
& t\'{e}l\'{e}phonie mobile s' ils sont utilis\'{e}s \`{a} bord '' , a d\'{e}clar\'{e} Rosenker .\\
\hline\hline
{\bf Our model} & 
Avec la cr\'{e}mation , il y a un `` sentiment de violence contre le corps d' un \^{e}tre cher '' , \\
& qui sera `` r\'{e}duit \`{a} une pile de cendres '' en tr\`{e}s peu de temps au lieu d' un processus de \\
& d\'{e}composition ``  qui accompagnera les \'{e}tapes du deuil '' .\\
\hline
{\bf Truth} & 
Il y a , avec la cr\'{e}mation , `` une violence faite au corps aim\'{e} '' , \\
& qui va \^{e}tre `` r\'{e}duit \`{a} un tas de cendres '' en tr\`{e}s peu de temps , et non apr\`{e}s un processus de \\
&d\'{e}composition , qui `` accompagnerait les phases du deuil '' .\\
\hline
\end{tabular}
\end{footnotesize}
\caption{A few examples of long translations produced by the LSTM
  alongside the ground truth translations.  The reader can verify that
  the translations are sensible using Google translate.  }
\label{tab:examples}
\end{table}

\subsection{Model Analysis}

\begin{figure}[h!]
\centering
\includegraphics[width=0.46\textwidth]{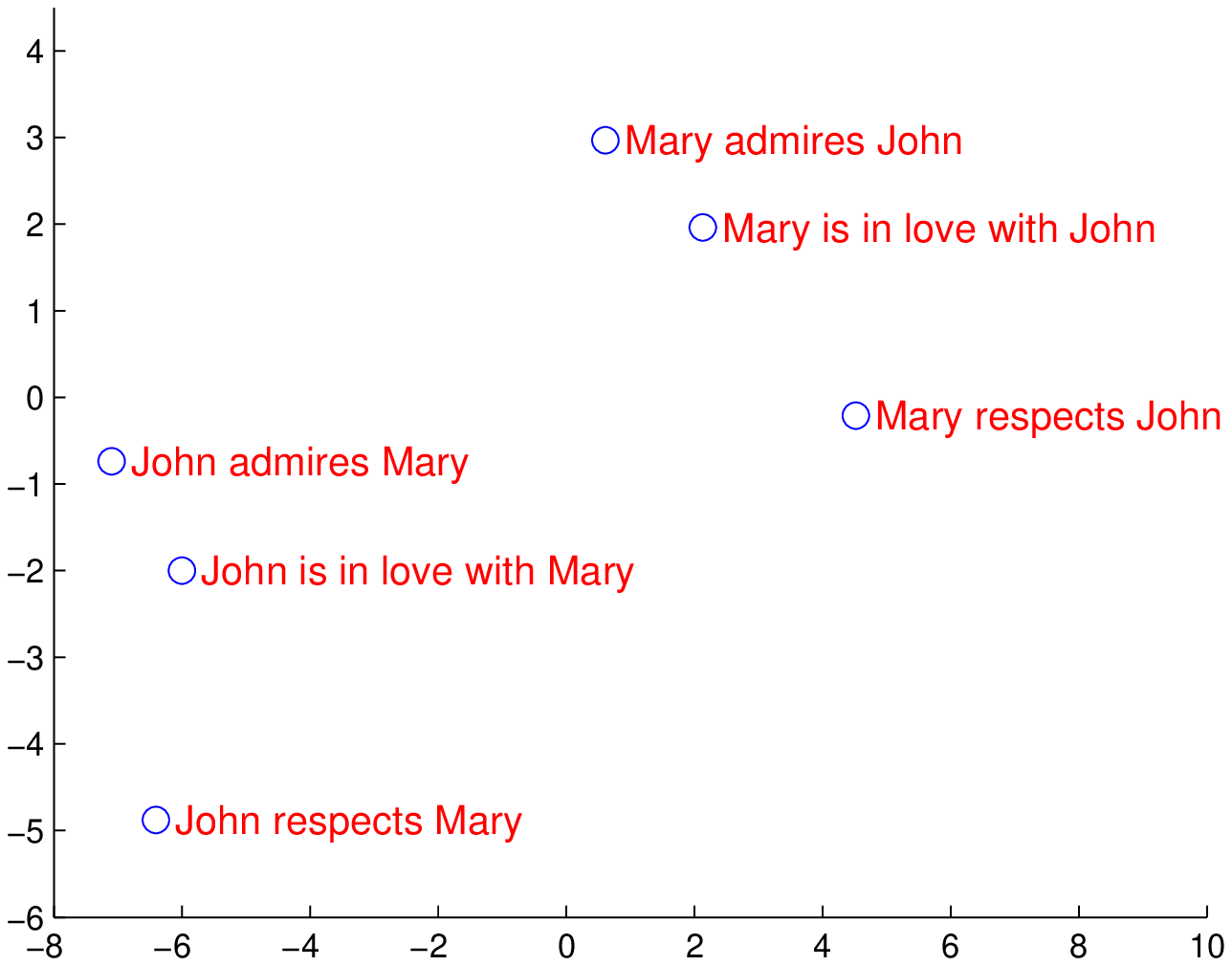} ~~
\includegraphics[width=0.46\textwidth]{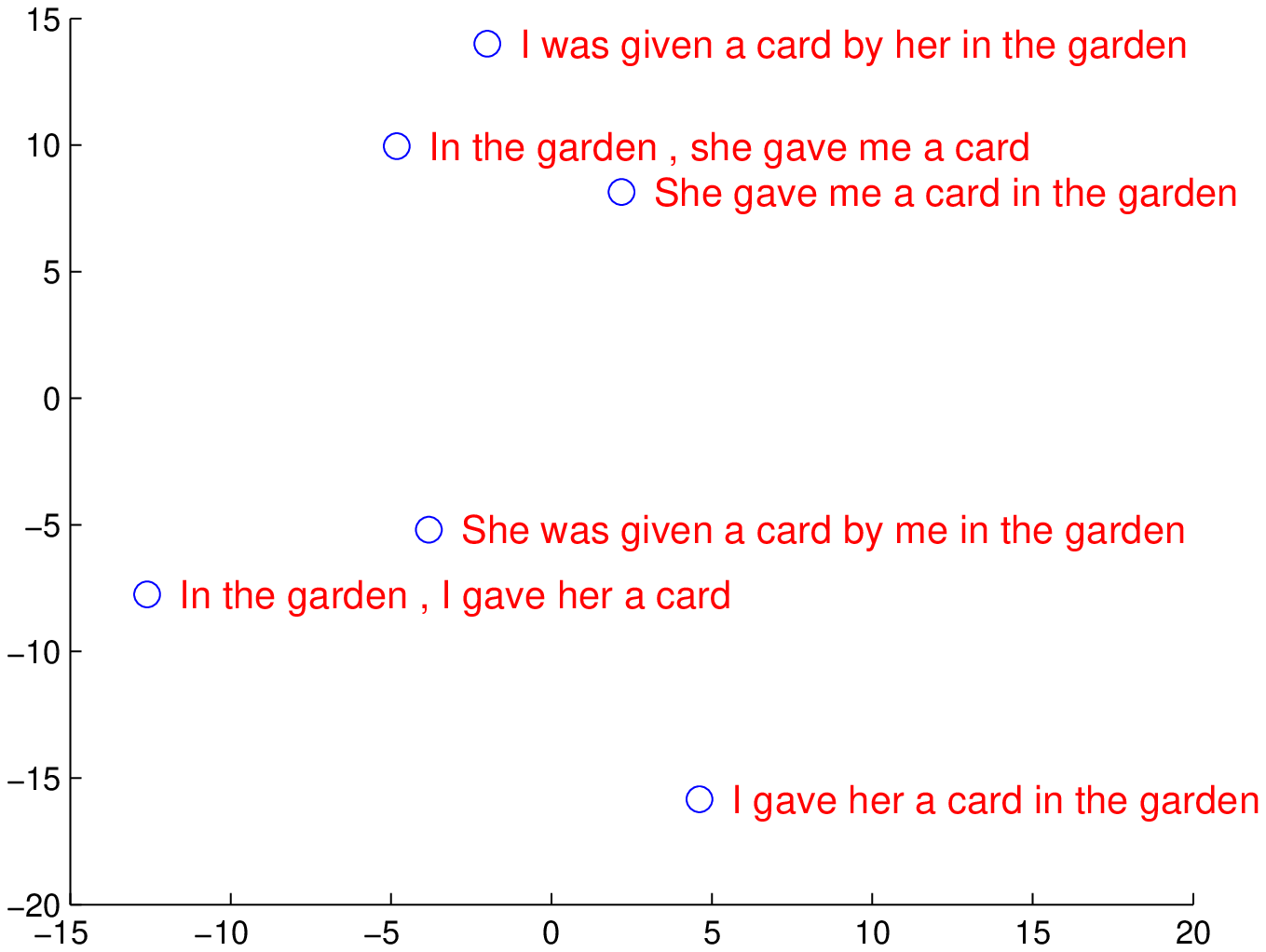} 
\caption{\small The figure shows a 2-dimensional PCA projection of the
  LSTM hidden states that are obtained after processing the phrases in
  the figures.  The phrases are clustered by meaning, which in these
  examples is primarily a function of word order, which would be
  difficult to capture with a bag-of-words model. Notice that both
  clusters have similar internal structure.}
\label{fig:embedding}
\end{figure}

One of the attractive features of our model is its ability to turn a
sequence of words into a vector of fixed dimensionality.
Figure~\ref{fig:embedding} visualizes some of the learned
representations.  The figure clearly shows that the representations
are sensitive to the order of words, while being fairly insensitive to
the replacement of an active voice with a passive voice.  The
two-dimensional projections are obtained using PCA.

\section{Related work}
\label{sec:rel_work}

There is a large body of work on applications of neural networks to
machine translation. So far, the simplest and most effective way of
applying an RNN-Language Model (RNNLM) \cite{mikolov2010recurrent} or
a Feedforward Neural Network Language Model (NNLM) \cite{bengio} to an
MT task is by rescoring the n-best lists of a strong MT baseline
\cite{mikolov2012}, which reliably improves translation quality.

More recently, researchers have begun to look into ways of including
information about the source language into the NNLM.  Examples of this
work include Auli et al.~\cite{auli13}, who combine an NNLM with a
topic model of the input sentence, which improves rescoring
performance.  Devlin et al.~\cite{devlin14} followed a similar
approach, but they incorporated their NNLM into the decoder of an MT
system and used the decoder's alignment information to provide the
NNLM with the most useful words in the input sentence.  Their approach
was highly successful and it achieved large improvements over their
baseline.

Our work is closely related to Kalchbrenner and Blunsom \cite{kal13},
who were the first to map the input sentence into a vector and then
back to a sentence, although they map sentences to vectors using
convolutional neural networks, which lose the ordering of the words.
Similarly to this work, Cho et al.~\cite{cho14} used an LSTM-like RNN
architecture to map sentences into vectors and back, although their
primary focus was on integrating their neural network into an SMT
system.  Bahdanau et al.~\cite{bog14} also attempted direct
translations with a neural network that used an attention mechanism to
overcome the poor performance on long sentences experienced by Cho et
al.~\cite{cho14} and achieved encouraging results.  Likewise,
Pouget-Abadie et al.~\cite{curse} attempted to address the memory
problem of Cho et al.~\cite{cho14} by translating pieces of the source
sentence in way that produces smooth translations, which is similar to
a phrase-based approach.  We suspect that they could achieve similar
improvements by simply training their networks on reversed source
sentences.

End-to-end training is also the focus of Hermann et
al.~\cite{hermann14}, whose model represents the inputs and outputs by
feedforward networks, and map them to similar points in
space. However, their approach cannot generate translations directly:
to get a translation, they need to do a look up for closest vector in
the pre-computed database of sentences, or to rescore a sentence.

\section{Conclusion}

In this work, we showed that a large deep LSTM, that has a limited 
vocabulary and that makes almost no
assumption about problem structure can outperform a standard SMT-based system whose vocabulary
is unlimited on a large-scale MT task.  The success of our simple
LSTM-based approach on MT suggests that it should do well on many
other sequence learning problems, provided they have enough training
data.

We were surprised by the extent of the improvement obtained by
reversing the words in the source sentences.  We conclude that it is
important to find a problem encoding that has the greatest number of
short term dependencies, as they make the learning problem much
simpler.  In particular, while we were unable to train a standard
RNN on the non-reversed translation problem (shown in
fig.~\ref{fig:translation-model2}), we believe that a standard RNN
should be easily trainable when the source sentences are reversed (although we
did not verify it experimentally).

We were also surprised by the ability of the LSTM to correctly
translate very long sentences.  We were initially convinced that the
LSTM would fail on long sentences due to its limited memory, and other
researchers reported poor performance on long sentences with a model
similar to ours \cite{cho14,bog14,curse}.  And yet,
LSTMs trained on the reversed dataset had little difficulty translating long
sentences.

Most importantly, we demonstrated that a simple, straightforward and a
relatively unoptimized approach can outperform an SMT system, so
further work will likely lead to even greater translation accuracies.  
These results suggest that our approach will likely   
do well on other challenging sequence to sequence problems.

\small
\section{Acknowledgments}

We thank Samy Bengio, Jeff Dean, Matthieu Devin, Geoffrey Hinton, Nal Kalchbrenner, Thang Luong, Wolfgang
Macherey, Rajat Monga, Vincent Vanhoucke, Peng Xu, Wojciech Zaremba,
and the Google Brain team for useful comments and discussions.

\bibliography{translate} 
\bibliographystyle{plain}
\end{document}